\documentclass[a4paper,10pt]{article}
\usepackage[utf8]{inputenc}
\usepackage{graphicx}
\usepackage{makecell}
\usepackage{float}
\usepackage{booktabs}
\usepackage{amsmath}
\usepackage{caption}
\usepackage{multicol}
\usepackage{geometry}
\usepackage{ragged2e} 
\usepackage{titling} 
\usepackage{abstract} 

\title{\textbf{Melon Fruit Detection and Quality Assessment Using Generative AI-Based Image Data Augmentation}}
\author{
  \begin{minipage}{\textwidth}
  \centering
  Seungri Yoon\textsuperscript{$\dagger$}\textsuperscript{1}, 
  Yunseong Cho\textsuperscript{$\dagger$}\textsuperscript{2}, 
  Tae In Ahn\textsuperscript{$\ast$}\textsuperscript{3,4} \\
  \textsuperscript{1}\small Graduate student, Department of Agriculture, Forestry and Bioresources (Horticultural Science and Biotechnology), Seoul National University, Seoul 08826, Korea \\
  \textsuperscript{2}\small Graduate student, Department of Mathematics, Mathematical Data Science, Korea University, Seoul 02841, Korea \\
  \textsuperscript{3}\small Assistant Professor, Department of Agriculture, Forestry and Bioresources (Horticultural Science and Biotechnology), Seoul National University, Seoul 08826, Korea \\
  \textsuperscript{4}\small Assistant Professor, Research Institute of Agriculture and Life Sciences, Seoul National University, Seoul 08826, Korea \end{minipage}
}

\date{}
\begin{document}

\maketitle

\renewcommand{\thefootnote}{\fnsymbol{footnote}}
\footnotetext[2]{Equal contributions.}
\footnotetext[1]{Corresponding author: tiahn@snu.ac.kr}

\begin{abstract}
\begin{justify}
Monitoring and managing the growth and quality of fruits are very important tasks. To effectively train deep learning models like YOLO for real-time fruit detection, high-quality image datasets are essential. However, such datasets are often lacking in agriculture. Generative AI models can help create high-quality images. In this study, we used MidJourney and Firefly tools to generate images of melon greenhouses and post-harvest fruits through text-to-image, pre-harvest image-to-image, and post-harvest image-to-image methods. We evaluated these AI-generated images using PSNR and SSIM metrics and tested the detection performance of the YOLOv9 model. We also assessed the net quality of real and generated fruits. Our results showed that generative AI could produce images very similar to real ones, especially for post-harvest fruits. The YOLOv9 model detected the generated images well, and the net quality was also measurable. This shows that generative AI can create realistic images useful for fruit detection and quality assessment, indicating its great potential in agriculture. This study highlights the potential of AI-generated images for data augmentation in melon fruit detection and quality assessment and envisions a positive future for generative AI applications in agriculture.
\end{justify}
\end{abstract}

\vspace{2em} 

\begin{multicols}{2}

\section{Introduction}

\indent In recent years, rapid advances in artificial intelligence (AI) and deep learning technologies have significantly improved the precision and efficiency of fruit detection and quality assessment \cite{lecun2015, bengio2021}. Specifically, the deep learning-based object detection algorithm YOLO (You Only Look Once) has demonstrated high performance in real-time detection, attracting significant attention from researchers \cite{redmon2016, moon2021}. Training these deep learning models effectively requires large-scale, high-quality image datasets \cite{zhao2019}. However, in agriculture, there is a serious lack of such large-scale image datasets. This shortage is due to the significant time and resources needed to create datasets that cover crops under various conditions and environments \cite{lu2020}.
To address this issue, AI-generated images have recently gained attention \cite{meoryahaya2023}. Generative AI models can produce high-quality images through text-to-image, image-to-image, and image-to-video transformations \cite{karras2019}. This technology has great potential to address the problem of limited large-scale image datasets through data augmentation \cite{shorten2019}.

Synthetic images created by computer algorithms have been widely used in various fields, including healthcare, fashion, architecture, geospatial studies, and agriculture \cite{chen2021a, bermano2022, choi2022, luo2022, sapkota2022}. Traditional image generation methods, such as parametric techniques \cite{chen2021b}, ray tracing \cite{mildenhall2022}, and physics-based rendering \cite{hodan2019}, have advanced synthetic image technology. However, each of these methods has limitations: they struggle to adapt to complex shapes \cite{velikina2013}, have high computational demands and time issues, and lack flexibility \cite{eversberg2021, diolatzis2022, sapkota2024}. In contrast, generative AI offers substantial benefits for research and development by allowing extensive variations and information to be added to datasets with significantly less time and cost compared to traditional data augmentation techniques \cite{wong2016, shorten2019, khalifa2022}. Previous studies on synthetic data enhancement for tasks such as tomato leaf disease classification, pest synthetic image generation, and weed classification \cite{lu2019, abbas2021, chen2024} show that generative AI is a valuable tool for improving the quality of agricultural data sets.

Muskmelon, prized for its attractive appearance and flavor, has a high market value. Throughout its growth from fruit set stage to harvest, the fruit undergoes significant physiological and morphological changes \cite{kano2006, ezura2009}. During the fruit enlargement stage, rapid growth causes the skin to harden and crack, forming the net pattern, a crucial factor in the commercial value of the melon. To enhance the external quality of muskmelons, it is necessary to analyze the fruit development stages, monitor growth, and manage quality, although this is a challenging task. In this context, fruit detection and quality assessment are crucial research topics in agriculture, significantly impacting crop monitoring, harvesting, quality control, and sorting automation \cite{wang2022}.

This study aims to explore the potential of generative images for data augmentation by applying and evaluating methodologies for fruit detection and quality assessment. To achieve this, the quality of the generative images was evaluated, melons were detected in the generative images using YOLOv9 fine-tuned with the melon fruit images, and the net evaluation method \cite{yoon2023} was applied to assess the external quality of the fruit. Through this, our objective is to confirm the potential of generative images as high-quality data in agriculture and to investigate their practical use.

\section{Materials and Methods}

\subsection{Cultivation and data collection}

The cultivation experiment was conducted in a Venlo-type glass greenhouse at the Protected Horticulture Research Institute (NIHHS, Haman, Korea). The muskmelon variety 'Dalgona' (Cucumis melo L., cv. Dalgona) was sown on May 9, 2023, using groundwater and subsequently transplanted into coir substrates (100×20×10 cm, Duckyang Agrotech, Nonsan, Korea) after three weeks. The Yamazaki standard nutrient solution suitable for melons was supplied at electrical conductivity levels of 1.8, 2.0, and 2.3 dSm\textsuperscript{-1} during the early, middle and late stages of growth, respectively. The drainage ratios for each stage were 40-50\%, 30-40\%, and 5-10\%. Daytime and nighttime temperatures were controlled at 33°C and 19°C. Pollination was carried out by bees on May 30, 2023, and fruit thinning was performed on June 5, 2023. After the fruit set between the 11\textsuperscript{th} and 13\textsuperscript{th} nodes, the fruit was thinned to leave one fruit per plant, and the plant was pruned at the 23\textsuperscript{rd} node to maintain a single stem.

The images of the melon fruits were obtained before and after harvest (Fig.1A). The pre-harvest images were taken using a smartphone (Galaxy S22, Samsung Electronics Inc., Suwon, Korea) with a resolution of 1920 × 1080 pixels under natural light conditions. The camera was positioned to capture the entire fruit from below to provide a comprehensive view. Post-harvest images were collected in a square-shaped studio (W80 × L80 × H80 cm) with LED lighting at the top, using the same camera.
\subsection{Generative AI}
To create melon fruit images, we used two popular generative AI tools (Fig.1B): MidJourney (MidJourney Basic Plan, MidJourney, San Francisco, CA, US) and Firefly (Fi

\end{multicols}

\begin{figure}[H]
    \centering
    \includegraphics[width=\textwidth]{"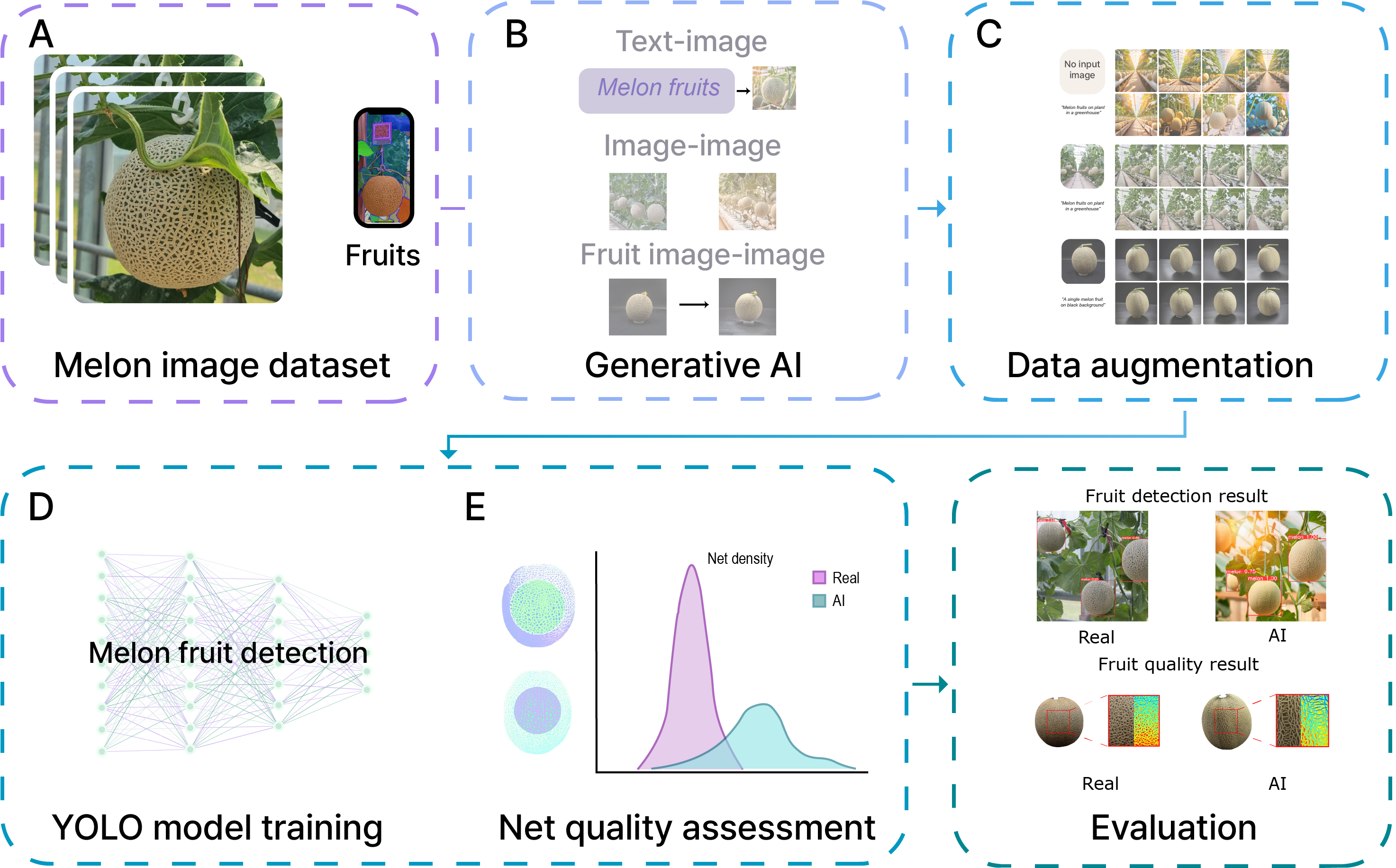"}
    \caption{Overview of the experiment. Process of melon image data collection through cultivation experiments \textbf{(A)}, three stages of image generation using generative AI \textbf{(B)}, data augmentation and evaluation \textbf{(C)}, YOLO model training \textbf{(D)}, net quality assessment \textbf{(E)}, and comprehensive evaluation.}
    \label{fig:1}
\end{figure}

\begin{multicols}{2}

\noindent
refly, Adobe, San Jose, CA, US). While text-to-image generation through prompts is useful for intuitive control, it has spatial ambiguity and does not allow sufficient user control. In this study, we generated images in three steps to capture changes in image quality and complex agricultural systems.

\subsubsection{Text-to-image generation}
Text-to-image generation involves carefully capturing the variations in the image through the topic, style, and parameters of the prompt. In this study, we used the 'describe' command in MidJourney to generate prompts from images and then created images based on these prompts. The images used for the prompts included greenhouse images of melon cultivation and individual fruit images after harvest.

\subsubsection{Image-to-image generation (Pre-harvest)}
Compared to text-to-image generation, the image-to-image generation technique references the structure and style of the original image while enhancing clarity and quality, producing accurate and realistic images \cite{sapkota2024}. Greenhouse images of melon cultivation were used, encompassing various stages of fruit growth, camera angles, lighting, and focus. These diverse images were used as reference images to generate similar structured images.

\subsubsection{Image-to-image generation (Post-harvest)}
Melon fruit images focused on individual fruits after harvest were generated for the purpose of evaluating the external quality of the melons.

\subsection{YOLOv9}
YOLOv9 is the latest version of real-time object detection technology, achieving significant advancements. Released in February 2024, this version introduces innovative technologies such as PGI (Programmable Gradient Information) and GELAN (Generalized Efficient Layer Aggregation Network). PGI addresses the problem of information being diluted or lost during the forward pass of the neural network, while GELAN emphasizes lightweight design, fast inference, and accuracy, directly addressing information bottlenecks \cite{vo2024}. YOLOv9 outperforms current state-of-the-art models in various metrics, maintaining equal or higher accuracy with fewer parameters. In this study, we trained YOLOv9 using a dataset of 700 real melon images and tested its detection performance on AI-generated images representing various environmental conditions (Fig.1D).

\subsection{Evaluation methods}

\subsubsection{Evaluation of AI-generated images}
Peak Signal to Noise Ratio (PSNR): PSNR is a metric used to evaluate image quality by measuring the ratio of the maximum potential power of the signal (represented by the original image) to the power of disruptive noise (represented by disparities between original and generated images). This ratio is calculated according to Equation \ref{eq:psnr}, and the mean squared error (MSE) is estimated according to Equation \ref{eq:mse}. Higher PSNR values indicate that the generated image is closer to the original image with minimal distortion \cite{sapkota2024}. Here, MAX\textsubscript{I} represents the maximum pixel value, and MSE indicates the pixel-level difference between the generated image (A) and the original image (O), with lower MSE values implying smaller differences between the images.

\begin{equation}
    \text{PSNR} = 10 \log_{10} \left(\frac{\text{MAX}_I^2}{\text{MSE}}\right)
    \label{eq:psnr}
\end{equation}

\begin{equation}
    \text{MSE} = \frac{1}{N} \sum_{i=1}^{N} (O - A)^2
    \label{eq:mse}
\end{equation}

Structural Similarity Index Measure (SSIM): SSIM is a metric for evaluating the quality of modified or generated images, similar to PSNR. It assesses the correlation between two images (x, y) based on three aspects: Luminance, Contrast, and Structure \cite{wang2004}. Luminance measures the brightness of light, contrast measures the difference in brightness, and structure measures the correlation. These values are calculated using the mean, standard deviation, and covariance of the pixels in the images.

\begin{equation}
    \text{SSIM}(x,y) = [l(x,y)]^\alpha \cdot [c(x,y)]^\beta \cdot [s(x,y)]^\gamma
    \label{eq:ssim}
\end{equation}

\subsubsection{Accuracy of YOLOv9 for fruit detection}

Intersection over Union (IoU): IoU is a metric that is used to evaluate the accuracy of object detection, to determine whether the detection of individual objects is successful. It has a value between 0 and 1. In computer vision and object detection, bounding boxes are typically represented as rectangles in 2D images. Based on this representation, the IoU between the actual bounding box (B\textsubscript{g}, Ground truth) and the predicted bounding box (B\textsubscript{d}) is calculated as follows \cite{zhou2019}:

\begin{equation}
    \text{IoU} = \frac{\text{Area\ of\ overlap\ between}~B_g~\text{and}~B_d}{\text{Area\ of\ union\ between}~B_g~\text{and}~B_d}
    \label{eq:iou}
\end{equation}

\subsubsection{Evaluation of net quality in generated images}

Net quality is a major evaluation metric in human visual inspection, evaluated based on the density and uniformity of the net pattern (Fig.1E). To visualize the differences in net quality, 20 fruits with high net quality were selected, and 20 generated images were classified. In this study, to distinguish it from the net, the non-netted exocarp region is referred to as skin. The masks containing only the fruit were then extracted from each dataset. Using a computer vision algorithm, the color of the skin and net were distinguished, and the average area (net density) and standard deviation (net uniformity) of the skin fragments in pixels were calculated \cite{yoon2023}.

\section{Results and Discussion}

\subsection{Evaluation of AI-generated images}
\begin{figure}[H] 
    \centering
    \includegraphics[width=\linewidth]{"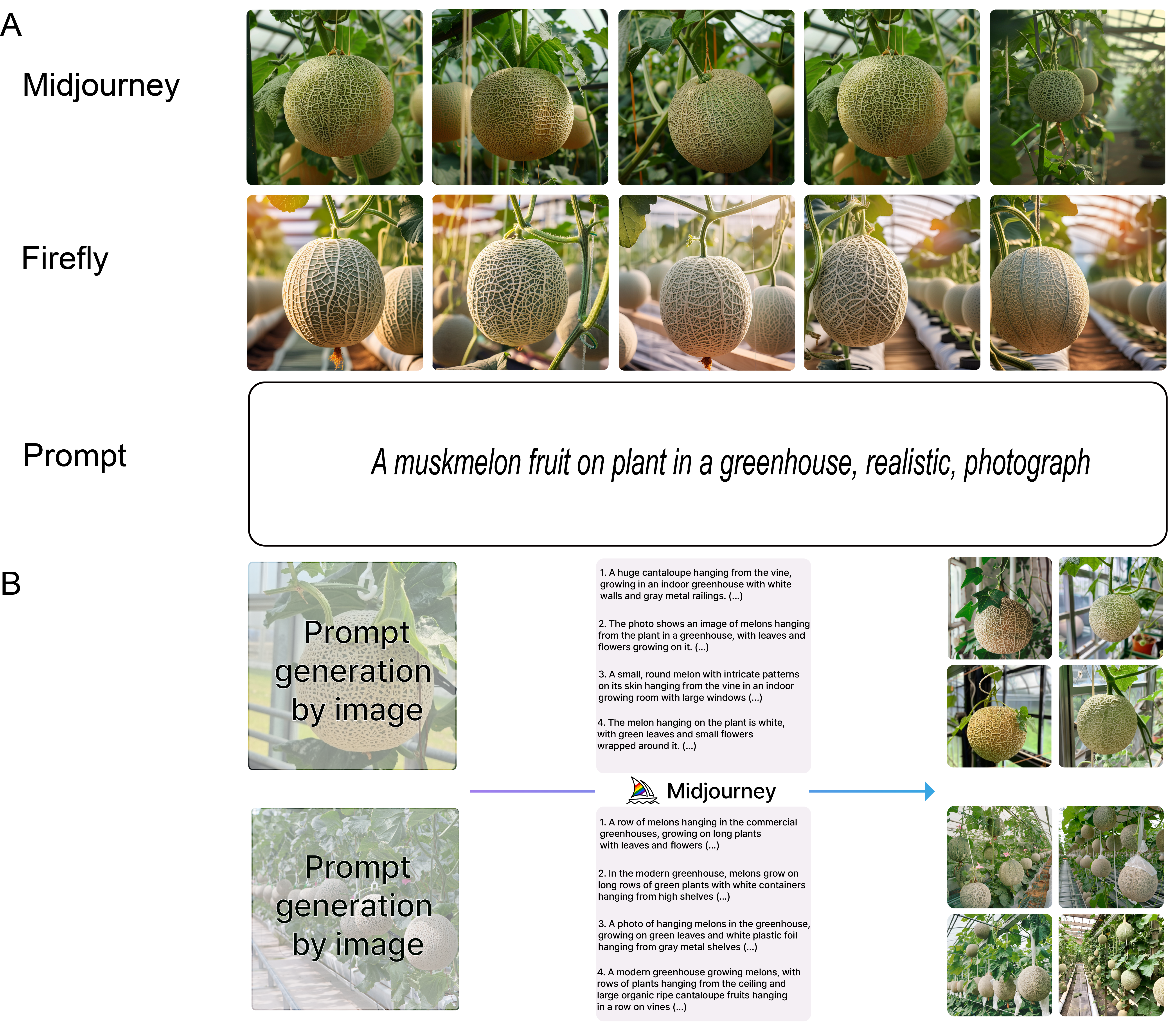"} 
    \caption{Text-to-image generation using Midjourney and Firefly (A), and the process of prompt generation from images followed by text-to-image generation (B)}
    \label{fig:2}
\end{figure}

The results of generating net melon images through text prompts are shown in Fig.2A. The two generative AI tools, Midjourney and Firefly, created images focusing on melons grown in greenhouses. However, when attempting to generate net melons (muskmelon variety) using only prompts without reference images, some images of cantaloupe with distinct green lines on the skin were generated. To produce high-quality images, prompt engineering with various word and parameter combinations, along with trial and error, is necessary. In this study, to reduce image variation from prompts, we used Midjourney’s prompt generation command (Fig.2B). This command allows the AI to analyze the original image and output semantic features as prompts. As a result, we obtained various prompts, including objects, backgrounds, camera angles and compositions, and styles, which helped us to create a new image set.
    
Image-to-image generation refers to a method that enhances the clarity and realism of the original image by referencing the structure and style of a source image. The results generated by referencing pre-harvest fruit images are very realistic, including images from various angles and lighting conditions (Fig.3). This method accurately reflected the detailed characteristics of hydroponic melon cultivation, including planting and training methods and the features of lateral branches, effectively recreating the actual greenhouse environment. Meanwhile, the post-harvest fruit image results detailed the external features of the melon, indicating that generative AI can be useful for evaluating the external quality of fruits (Fig.4).
\begin{figure}[!htbp]
    \centering
    \includegraphics[width=\linewidth]{"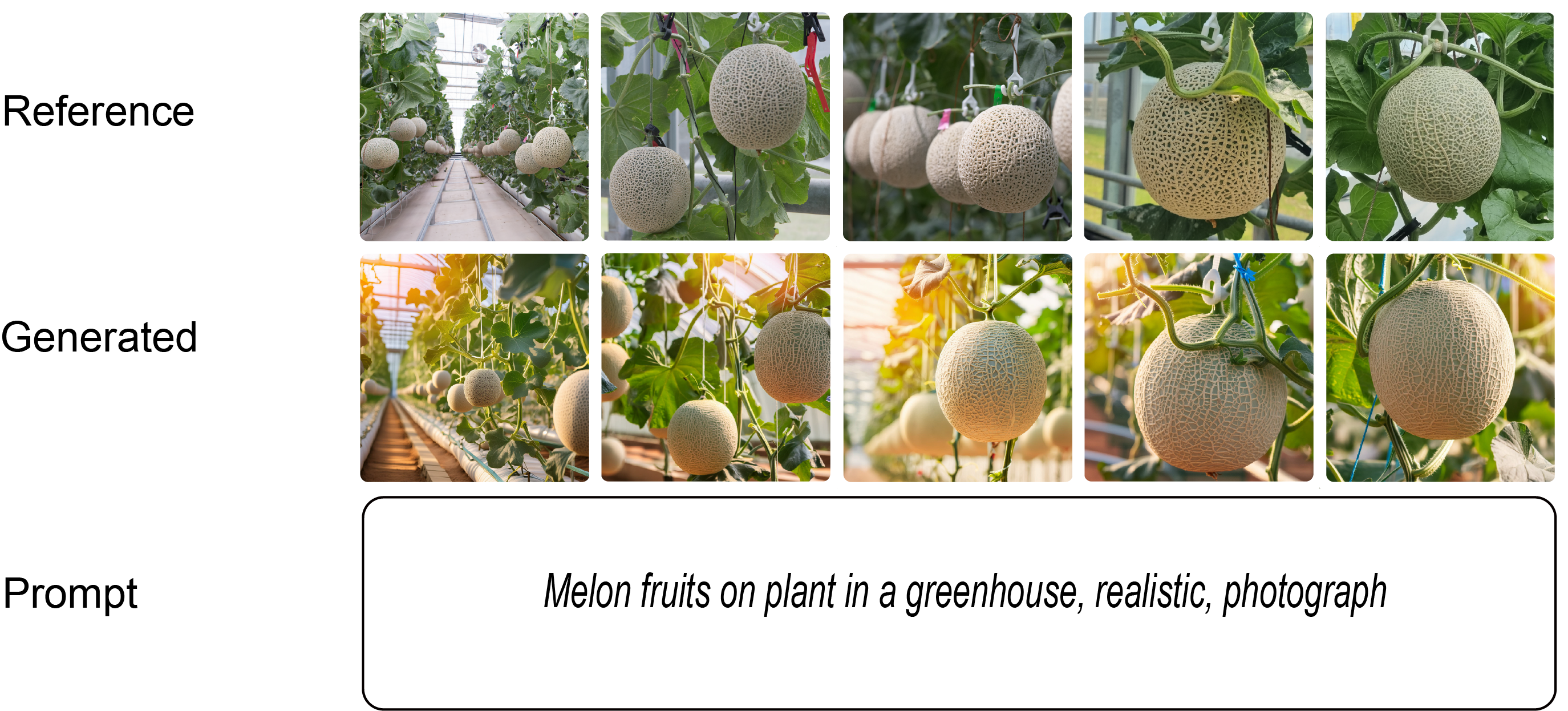"}
    \caption{Image-to-image generation process using a greenhouse with melon plants before harvest as reference}
    \label{fig:3}
\end{figure}

In this study, we augmented data using the three-step image generation methods described earlier, and we evaluated the quantitative similarity between the generated datasets and the original images (Fig.5). The evaluation results showed that the average and standard deviation of PSNR values for text-to-image, image-to-image (pre-harvest), and image-to-image (post-harvest) dataset groups were 27.4±0.6, 27.9±0.05, and 28.8±0.3, respectively (Table 1). The SSIM values, which indicate a higher similarity to the original image when closer to 1, were 0.06±0.02, 0.12±0.03, and 0.42±0.1. These results suggest that images generated through text and pre-harvest references showed a high similarity to the original images, but also had significant changes in brightness, contrast, and structure. In contrast, data augmentation using post-harvest individual fruit images resulted in high similarity to the original images, while also capturing structural features accurately due to the limited background and focused fruit content.

\begin{figure}[H] 
    \centering
    \includegraphics[width=\linewidth]{"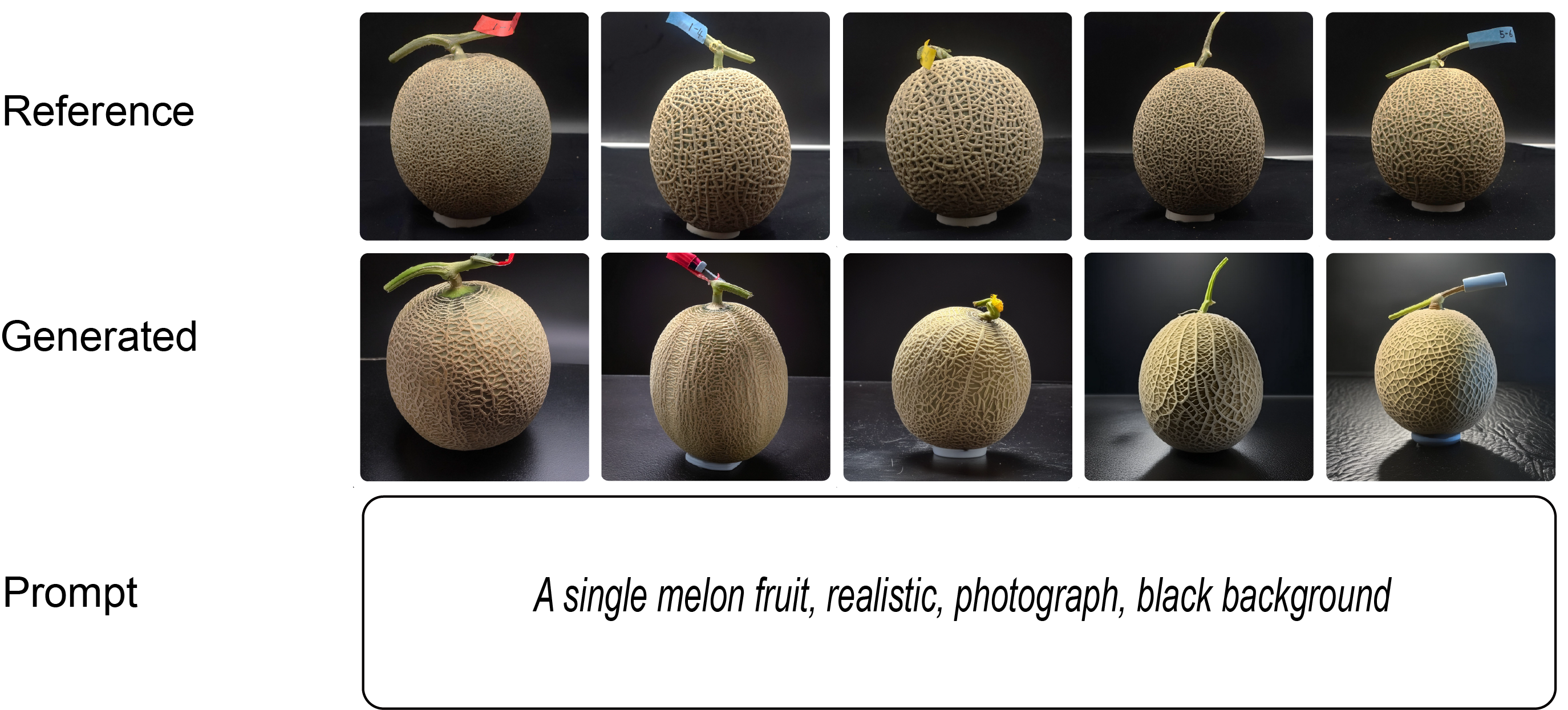"} 
    \caption{Image-to-image generation process using single melon fruit images after harvest as reference}
    \label{fig:4}
\end{figure}

\begin{figure}[H] 
    \centering
    \includegraphics[width=\linewidth]{"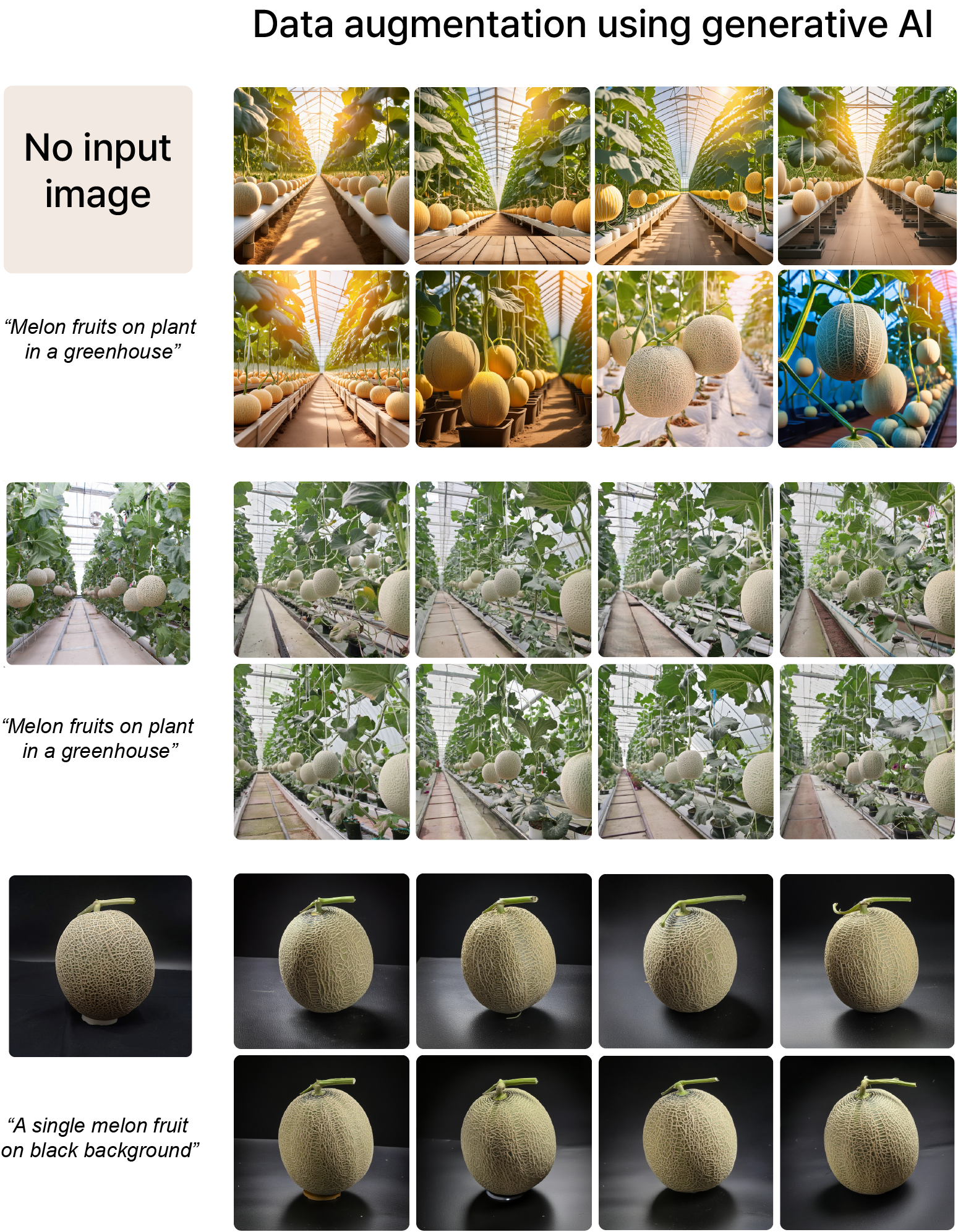"} 
    \caption{Data augmentation using generative AI across three image generation processes}
    \label{fig:5}
\end{figure}

\end{multicols}

\begin{table}[H]
    \centering
    \caption{Evaluation metrics for PSNR and SSIM across different AI-generated image conditions. The PSNR and SSIM values indicate the quality and similarity of the images generated under each condition. Different letters (a, b, c) denote statistically significant differences between the groups at $p < 0.05$. The asterisks (***) represent the level of statistical significance determined by the Tukey HSD post-hoc test ($p < 0.001$).}
    \label{tab:psnr_ssim}
    \begin{tabular}{p{3cm}p{3cm}p{3cm}p{3cm}}
        \hline
        Metric & Text-image & Image-image (pre-harvest) & Image-image (post-harvest) \\
        \hline
        PSNR & 27.5 c & 27.9 b & 28.8 a \\
        SSIM & 0.06 c & 0.12 b & 0.42 a \\
        Significance & *** & *** & *** \\
        \hline
    \end{tabular}
\end{table}

\begin{multicols}{2}

\subsection{Fruit detection using AI-generated images}
\begin{figure}[H] 
    \centering
    \includegraphics[width=\linewidth]{"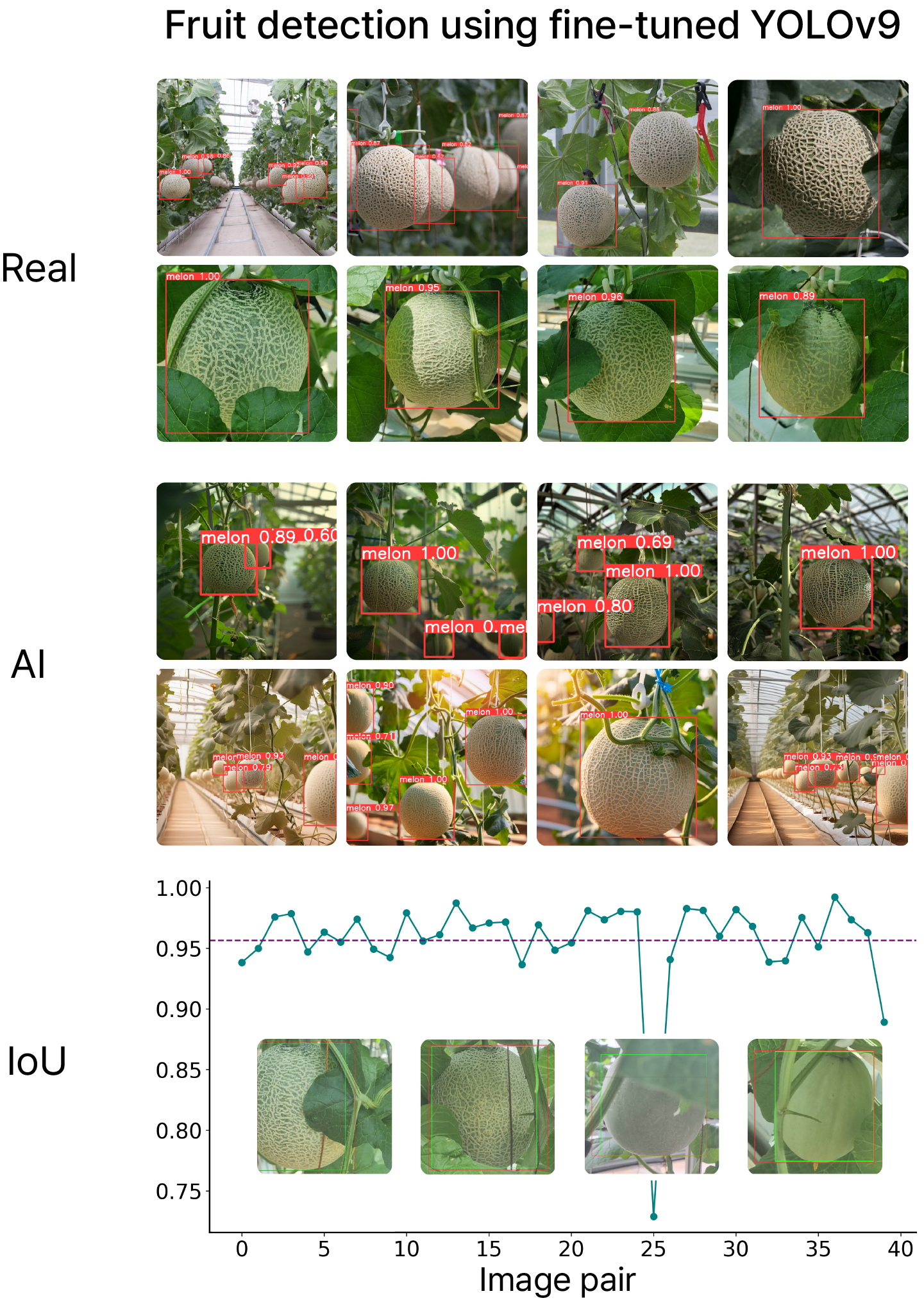"} 
    \caption{Fruit detection using fine-tuned YOLOv9 on real images and AI-generated images of melons and the Intersection over Union (IoU) evaluation. Example images within the IoU plot show the ground truth bounding boxes (green) images and predicted bounding boxes (red).}
    \label{fig:6}
\end{figure}
The results of detecting melons at various growth stages using the proposed YOLOv9 image processing algorithm are shown in Fig.6. YOLOv9 achieved a very high performance with an IoU of 0.95 for melon fruit detection using 40 real bounding boxes (Ground truth). This high performance was consistent across various growth stages in the greenhouse, including fruit enlargement, net formation, and maturation, effectively detecting objects even when they were obscured by leaves. Additionally, YOLOv9 effectively detected melons in images generated by the generative AI tools Midjourney and Firefly, just as it did with real images. The generated images included a variety of colors, camera angles, and environmental conditions (e.g., sunny days, cloudy days).
The key factors influencing the detection and classification performance of deep learning models include the quality of image data, the image capture conditions, and the hyperparameter settings of each model \cite{wang2022}. However, agricultural environments pose challenges for collecting high-quality data due to dense planting, mutual shading by leaves and crops, intense lighting from sunlight, or shading by greenhouse structures. Therefore, overcoming false positives caused by these structural features of agricultural settings is a critical task for researchers. Large datasets that include a wide range of object classes, lighting conditions, and backgrounds help the model learn extensive features \cite{yu2015}. For example, common datasets like COCO, ImageNet, and Pascal VOC contain millions of annotated images, covering various categories and conditions \cite{wang2021}. In this context, generative AI offers the advantage of recreating diverse cultivation environments without the constraints of time and place, maximizing the quantity and diversity of data available for model development.

\subsection{Quality assessment of fruits using AI-generated images}

In this study, we applied a methodology to analyze the pores of muskmelons using computer vision technology and quantified net density and uniformity to assess quality \cite{yoon2023}. Fig. 7AB shows representative original images with valid masks and overlapping Regions of Interest (ROI), comparing the nets of high-quality real fruits with those of generated images. Through binarization techniques, we were able to highlight areas of the skin that contrasted with the net as white, although complete separation was challenging (Fig.7C). Using the depth estimation technique, we detected shallow and deep areas in 2D images, allowing us to distinguish between the net and the skin (Fig.7D). Small red-pixel patches of skin were considered individual islands, with lower individual area values indicating higher net density, and lower standard deviation of the area values indicating higher net uniformity (Fig.7E). This allowed us to quantify the net quality of both real and generated fruits (Fig.7F).
The nets in the AI-generated images detailed specific features of real fruit, such as bumps, cracks, and colors. Since the fruits used in the experiment were selected for their high net quality, both net density and uniformity were high (Fig.7F). The generated fruits showed lower net density and greater uniformity variation, resulting in lower quality evaluations compared to real fruits. However, it is encouraging that the AI-generated images were realistic enough to be used in actual net quality assessment methods.

\end{multicols}

\begin{figure}[!htb]
    \centering
    \includegraphics[width=0.9\textwidth]{"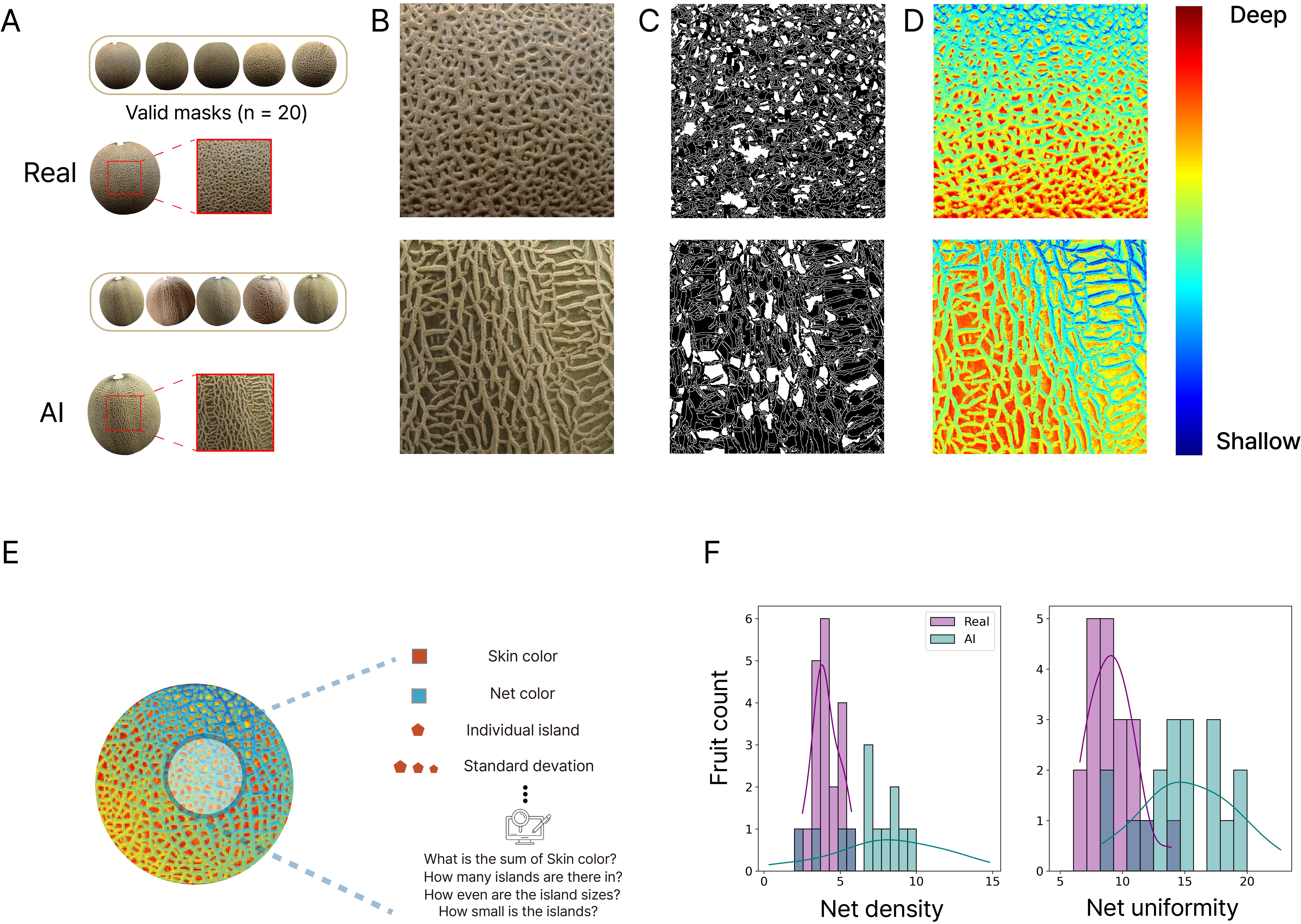"}
    \caption{Net quality assessment process for muskmelons. Steps for analyzing melon masks: Regions of interest (ROI) on real and AI-generated melons, including valid masks from both real and AI-generated melons (n=20) (A); Close-up views of the nets (B); Binarized images where white areas represent the skin area, showing that smaller and more uniform areas correspond to higher net quality (C); Depth estimation images, with red indicating the skin and blue indicating the net (D); Depiction of net quality quantification using computer vision (E): The skin color is differentiated from the net color, with each net fragment treated as a distinct island. Net density was calculated by dividing the total island area (skin area) by the number of islands; Quantification results of net density and uniformity, comparing real and AI-generated melons (F)}
    \label{fig:7}
\end{figure}

\begin{multicols}{2}

The net of a melon is a result of precise irrigation strategies during the growing period and is closely related to sweetness, making it a key indicator of market value \cite{leiva2013, lim2020}. Obtaining a variety of net images through generated images is expected to be effectively used in the development of fruit quality and grading models \cite{bird2022}.

\subsection{Limitations and implications}

In this study, we used text-to-image generation, pre-harvest image-to-image generation, and post-harvest image-to-image generation to obtain melon images simulating various environments. Text-to-image generation allows users to intuitively request desired images and easily express creative ideas \cite{dehouche2023}. However, it may be challenging to generate the desired images due to semantic ambiguity, text dependency, and limited interpretability. However, image-to-image generation produces new images based on reference images, providing specific and consistent results. Nevertheless, it has limitations in terms of transformation and creativity, is highly dependent on the quality of the original image, and the adjustment of detailed features can be more complex.

Despite these limitations, AI-generated images have shown great potential for research and development by detailing the melon hydroponic cultivation system and the external quality of post-harvest fruits. Beyond melon fruits, generative AI can be used to collect images of various crops at different stages of vegetative and reproductive growth, accurately depict surface conditions to capture physiological disorder images, and obtain images of fruits segmented by quality grades based on marketability. This means that high-quality AI-generated data can improve the performance of automated models related to growth monitoring, pest diagnosis, shipment, and distribution management, in addition to fruit detection and classification. 

With the rapid advancement of AI generation tools, the results of this study are expected to contribute to the application of generative AI in the fields of agriculture and plant science.

\section{Conclusion}
In this study, we used the tools Midjourney and Firefly to generate images of melon cultivation in greenhouses and post-harvest fruit through text-to-image, image-to-image (pre-harvest), and image-to-image (post-harvest) methods. The generated images were evaluated using metrics like PSNR and SSIM, and the detection performance of the YOLOv9 model (IoU = 0.95) was also assessed. In addition, we evaluated the net quality of real and generated fruits using a net quality assessment method. The generative AI produced images similar to the original, with post-harvest images showing exceptional realism. These generated images were detected by the YOLOv9 model as real images, and their net quality was also assessable. This shows that generative AI can create realistic images for practical and effective fruit detection and quality assessment, highlighting its significant potential in agriculture.

\end{multicols}
\end{document}